\documentclass{article}

\usepackage{iclr2018_conference, times} 

\usepackage{graphicx} 
\usepackage{subfig}
\usepackage{xspace}
\usepackage{caption}
\usepackage{color}
\usepackage{cprotect}

\usepackage{natbib}

\usepackage{algorithm}
\usepackage{algorithmic}

\usepackage{hyperref}

\usepackage{verbatim}		

\usepackage{url}		
\usepackage{fancyhdr}		
\usepackage{datetime}		
\usepackage{lastpage}		

\usepackage{amsmath}
\usepackage{amssymb}
\usepackage{amsthm}
\usepackage{xspace}

\newtheorem{theorem}{Theorem}[section]

\newtheorem{proposition}[theorem]{Proposition}

\newenvironment{myproof}[1][\proofname]{\proof[#1]\mbox{}}{\endproof}

\newcommand{\concat}{\text{concat}}
\newcommand{\logdensename}{Log-Dense Network\xspace}
\newcommand{\logdense}{Log-DenseNet\xspace}
\newcommand{\logdenses}{Log-DenseNets\xspace}
\newcommand{\logdensenp}{Log-DenseNet}

\newcommand{\loglogdense}{LogLog-DenseNet\xspace}

\newcommand{\nearest}{NEAREST\xspace}
\newcommand{\evenspace}{EVENLY-SPACED\xspace}
\newcommand{\pbdfull}{maximum backpropagation distance\xspace}
\newcommand{\pbd}{MBD\xspace}
\newcommand{\bd}{\text{BD}\xspace}

\newcommand{\round}[1]{\lfloor #1 \rceil}

\newcommand{\naive}{na\"{\i}ve\xspace}

\usepackage{tablefootnote}
\usepackage{scrextend}

\usepackage{enumitem} 
\setitemize{itemsep=0mm, leftmargin=5mm, topsep=0mm}
\setenumerate{itemsep=0mm, leftmargin=5mm, topsep=0mm}
\usepackage[noindentafter]{titlesec}
\titlespacing\section{0pt}{4pt plus 0pt minus 2pt}{2pt plus 0pt minus 1pt}
\titlespacing\subsection{0pt}{4pt plus 0pt minus 1pt}{1pt plus 1pt minus 1pt}
\titlespacing\subsubsection{0pt}{4pt plus 0pt minus 1pt}{1pt plus 1pt minus 1pt}

\title{\logdense: How to Sparsify a DenseNet}

\author{Hanzhang Hu$^1$, Debadeepta Dey$^2$, Allison Del Giorno$^1$, Martial Hebert$^1$ \& J. Andrew Bagnell$^1$ \\
\resizebox{\textwidth}{!}{
\begin{tabular}[h]{@{}l}
    $^1$ Carnegie Mellon University\\
    Pittsburgh, PA, USA \\
    \texttt{\{hanzhang,adelgior,hebert,dbagnell\}@cs.cmu.edu}
\end{tabular}
\hfill
\begin{tabular}[h]{l@{}}
    $^2$ Microsoft Research \\
    Redmond, WA, USA \\
    \texttt{dedey@microsoft.com}
\end{tabular}
}
}

\iclrfinalcopy
\begin{document}
\maketitle



\begin{abstract}
Skip connections are increasingly utilized by deep neural networks to improve accuracy and cost-efficiency. 
In particular, the recent DenseNet is efficient in computation and parameters, and achieves state-of-the-art predictions by directly connecting each feature layer to all previous ones. However, DenseNet's extreme connectivity pattern may hinder its scalability to high depths, and in applications like fully convolutional networks, full DenseNet connections are prohibitively expensive. 
This work first experimentally shows that one key advantage of skip connections is to have short distances among feature layers during backpropagation. Specifically, using a fixed number of skip connections, the connection patterns with shorter backpropagation distance among layers have more accurate predictions. Following this insight, we propose a connection template, \logdense, which, in comparison to DenseNet,  only slightly increases the backpropagation distances among layers from 1 to  ($1 + \log_2 L$), but uses only $L\log_2 L$ total connections instead of $O(L^2)$. Hence, \logdenses are easier than DenseNets to implement and to scale. We demonstrate the effectiveness of our design principle by showing better performance than DenseNets on \textit{tabula rasa} semantic segmentation, and competitive results on visual recognition.



\end{abstract}

\section{Introduction}

Deep neural networks have been improving performance for many machine learning tasks, scaling from networks like AlexNet~\citep{alexnet} to increasingly more complex and expensive networks, like VGG~\citep{vggnet}, ResNet~\citep{resnet} and Inception~\citep{inception_v4}.  Continued hardware and software advances will enable us to build deeper neural networks, which have higher representation power than shallower ones. 
However, the payoff from increasing the depth of the networks only holds in practice if the networks can be trained effectively. 
It has been shown that {\naive}ly scaling up the depth of networks actually decreases the performance \citep{resnet}, partially because of vanishing/exploding gradients in very deep networks. 
Furthermore, in certain tasks such as semantic segmentation, it is common to take a pre-trained network and fine-tune, because training from scratch is difficult in terms of both computational cost and reaching good solutions. Overcoming the vanishing gradient problem and being able to train from scratch are two active areas of research.

Recent works attempt to overcome these training difficulties in deeper networks by introducing skip, or shortcut, connections~\citep{fcn, hypercolumn, highwaynet, resnet, fractalnet, densenet} so the gradient reaches earlier layers and compositions of features at varying depth can be combined for better performance. In particular, DenseNet~\citep{densenet} is the extreme example of this, concatenating all previous layers to form the input of each layer, i.e., connecting each layer to all previous ones. However, this incurs an $O(L^2)$ run-time complexity for a depth $L$ network, and may hinder the scaling of networks. Specifically, in fully convolutional networks (FCNs), where the final feature maps have high resolution so that full DenseNet connections are prohibitively expensive, \cite{fcdense} propose to cut most of connections from the mid-depth. To combat the scaling issue, \cite{densenet} propose to halve the total channel size a number of times. Futhermore, \cite{slim_nn} cut 40\% of the channels in DenseNets while maintaining the accuracy, suggesting that much of the $O(L^2)$ computation is redundant. Therefore, it is both necessary and natural to consider a more efficient design principle for placing shortcut connections in deep neural networks.


In this work, we address the scaling issue of skip connections by answering the question: if we can only afford the computation of a limited number of skip connections and we believe the network needs to have at least a certain depth, where should the skip connections be placed?
We design experiments to show that with the same number of skip connections at each layer, the networks can have drastically different performance based on where the skip connections are. In particular, we summarize this result as the following design principle, which we formalize in Sec.~\ref{sec:pbd}:
\textbf{given a fixed number of shortcut connections to each feature layer, we should choose these shortcut connections to minimize the distance among layers during backpropagation.}

Following this principle, we design a network template, \logdense. In comparison to DenseNets at depth $L$, \logdenses cost only $L\log L$, instead of $O(L^2)$ run-time complexity. Furthermore, \logdenses only slightly increase the short distances among layers during backpropagation from 1 to $1 + \log L$. Hence, \logdenses can scale to deeper and wider networks, even without custom GPU memory managements that DenseNets require. In particular, we show that \logdenses outperform DenseNets on \textit{tabula rasa} semantic segmentation on CamVid~\citep{camvid-dataset}, while using only half of the parameters, and similar computation. \logdenses also achieve comparable performance to DenseNet with the same computations on visual recognition data-sets, including ILSVRC2012~\citep{ilsvrc}. In short, our contributions are as follows:

\begin{itemize}
    \item We experimentally support the design principle that with a fixed number of skip connections per layer, we should place them to minimize the distance among layers during backpropagation. 
    \item The proposed \logdenses achieve small $1 + \log_2 L$ between-layer distances using few connections ($L\log_2 L$), and hence, are scalable for deep networks and applications like FCNs.
    \item The proposed network outperforms DenseNet on CamVid for \textit{tabula rasa}  semantic segmentation, and achieves comparable performance on ILSVRC2012 for recognition.
\end{itemize}

\section{Background and Related Works}
\label{sec:background}

\textbf{Skip connections.}
The most popular approach to creating shortcuts is to directly add features from different layers together, with or without weights. Residual and Highway Networks~\citep{resnet, highwaynet} propose to sum the new feature map at each depth with the ones from skip connections, so that new features can be understood as fitting residual features of the earlier ones. FractalNet~\citep{fractalnet} explicitly constructs shortcut networks recursively and averages the outputs from the shortcuts. Such structures prevent deep networks from degrading from the shallow shortcuts via ``teacher-student" effects. \citep{dropoutnet} implicitly constructs skip connections by allowing entire layers to be dropout during training. 
DualPathNet~\citep{dualpathnet} combines the insights of DenseNet~\citep{densenet} and ResNet~\citep{resnet}, and utilizes both concatenation and summation of previous features. 



\textbf{Run-time Complexity and Memory of DenseNets.}
DenseNet~\citep{densenet} emphasizes the importance of compositional skip connections and it is computationally efficient (in accuracy per FLOP) compared to many of its predecessors.  One intuitive argument for the cost-efficiency of DenseNet is that the layers within DenseNet are directly connected to each other, so that all layers can pick up training signals easily, and adjust accordingly.  However, the quadratic complexity may prevent DenseNet from scale to deep and wide models. In fact at each downsampling, DenseNet applies block compression, which halves the number of channels in the concatenation of previous layers. DenseNet also opts not to double the output channel size of conv layers after downsampling, which divides the computational cost of each skip connection. 
These design choices enable DenseNets to be deep for image classification where final layers have low resolutions. 
However, final layers in FCNs for semantic segmentation have higher resolution than in classification. Hence, to fit models in the limited GPU memory, FC-DenseNets~\citep{fcdense} have to cut most of their skip connections from mid-depth layers. 
Furthermore, a \naive implementation of DenseNet requires $O(L^2)$ memory, because the inputs of the $L$ convolutions are individually stored, and they cost $O(L^2)$ memory in total.   
Though there exist $O(L)$ implementations via memory sharing among layers~\citep{densenet_torch}, they require custom GPU memory management, which is not supported in many existing packages. Hence, one may have to use custom implementations and recompile packages like Tensorflow and CNTK for memory efficient Densenets, e.g., it costs a thousand lines of C++ on Caffe~\citep{densenet_caffe}. 
Our work recognizes the contributions of DenseNet's architecture to utilize skip connections, and advocates for the efficient use of compositional skip connections to shorten the distances among feature layers during backpropagation. Our design principle can especially help applications like FC-DenseNet~\citep{fcdense} where the network is desired to be at least a certain depth, but only a limited number of shortcut connections can be formed. 

\textbf{Network Compression.} 
A wide array of works have proposed methods to compress networks by reducing redundancy and computational costs. \citep{linear_structure_in_cnn, compress4mobile, deep_roots} decompose the computation of convolutions at spatial and channel levels to reduce convolution complexity. \citep{distillation, deepreally} propose to train networks with smaller costs to mimic expensive ones. \citep{slim_nn} uses $L1$ regularization to cut 40\% of channels in DenseNet without losing accuracy. These methods, however, cannot help in applications that cannot fit the complex networks in GPUs in the first place. This work, instead of cutting connections arbitrarily or post-design, advocates a network design principle to place skip connections intelligently to minimize between-layer distances.

\section{From DenseNet to \logdense} 
\subsection{Preliminary on DenseNets}
\label{sec:densenet-preliminary}
Formally, we call the feature layers in a feed-forward convolutional network as $x_0, x_1,..., x_L$, where $x_0$ is the feature map from the initial convolution on the input image $x$, and each of the subsequent $x_i$ is from a transformation $f_i$ with parameter $\theta_i$ that takes input from a subset of $x_0,...,x_{i-1}$. In particular, the traditional feed-forward networks have $x_i = f_i(x_{i-1}; \theta_i)$, and the skip connections allow $f_i$ to utilize more than just $x_{i-1}$ for computing $x_i$. Following the trend of utilizing skip connections to previous layers \citep{resnet, highwaynet, fractalnet}, DenseNet \citep{densenet} proposes to form each feature layer $x_i$ using all previous features layers, i.e., 
\begin{align}
   x_i = f_i( \concat(\{ x_j : j = 0,..., i-1\} )\;; \; \theta_i),
\end{align}
where $\concat(\bullet)$ concatenates all features in its input collection along the feature channel dimension. Each $f_i$ is a bottleneck structure~\citep{densenet}, BN-ReLU-1x1conv-BN-ReLU-3x3conv, where the final conv produces $g$, the growth rate, number of channels, and the bottleneck 1x1 conv produces $4g$ channels of features out of the merged input features. DenseNet also organizes layers into $n_{block}$ number of blocks. Between two contiguous blocks, there is a 1x1conv-BN-ReLU, followed by an average pooling, to transform and downsample all previous features maps together to a coarser resolution. In practice, $n_{block} \leq 4$ in almost all state-of-the-art visual recognition architectures~\citep{resnet, inception_v4, densenet}. 
The direct connections among layers in DenseNet are argued to be the key reason why DenseNets enjoy high efficiency in parameter and computation to achieve the state-of-the-art predictions: the direct connections introduce implicit deep supervision~\citep{supervisednet} in intermediate layers, and reduce the vanishing/exploding gradient problem by enabling direct influence between any two feature layers.



\subsection{\pbdfull}
\label{sec:pbd}
We now formally define the proposed design principle that with the same number of connections, the distance between any two layers during backpropagation should be as small as possible. We consider each $x_i$ as a node in a graph, and the directed edge $(x_i, x_j)$ exists if $x_i$ takes direct input from $x_j$. The \textit{backpropagation distance} (\bd) from $x_i$ to $x_j$ $(i > j)$ is then the length of the shortest path from $x_i$ to $x_j$ on the graph. Then we define the \textit{\pbdfull} (\pbd) as the maximum \bd among all pairs $i > j$. Then DenseNet has a \pbd of 1, if we disregard transition layers. To reduce the $O(L^2)$ computation and memory footprint of DenseNet, we propose \logdense which increase \pbd slightly to $1+\log_2 L$ while using only $O(L \log L)$ connections and run-time complexity. Since the current practical networks have less than 2000 depths, the proposed method has a \pbd of at most 7 to 11.

\subsection{\logdense}
\label{sec:logdense}

\begin{figure}
    \centering
    \subfloat[DenseNet]{
        \includegraphics[width=0.22\textwidth]{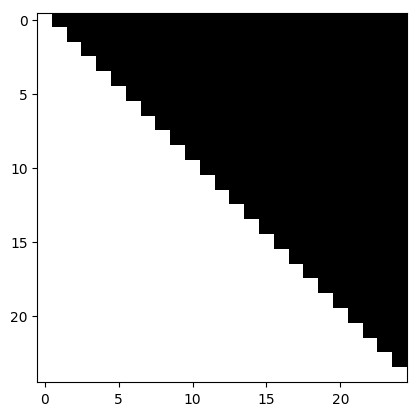}
        \label{fig:dense_connections}
    }
    ~
    \subfloat[\logdense V1]{
        \includegraphics[width=0.22\textwidth]{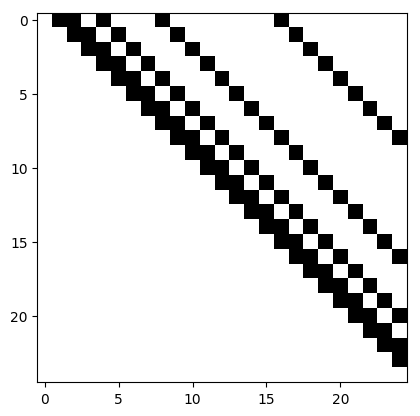}
        \label{fig:logv1_connections}
    }
    ~
    \subfloat[\logdense V2]{
        \includegraphics[width=0.22\textwidth]{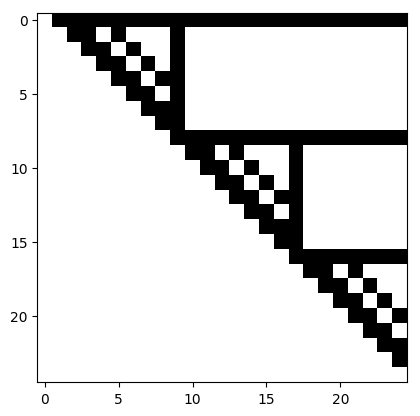}
        \label{fig:logv2_connections}
    }
    ~
    \subfloat[\loglogdense]{
        \includegraphics[width=0.22\textwidth]{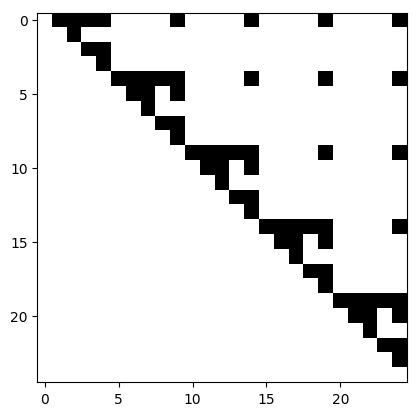}
        \label{fig:loglog_connections}
    }

    \cprotect\caption{Connection illustration for $L=24$. Layer 0 is the initial convolution.
    $(i,j)$ is black means $x_j$ takes input from $x_i$; it is white if otherwise. We assume there is a block transition at depth 12 for \logdense V2. \loglogdense is a connection strategy that has 2+$\log\log L$ \pbd.}
    \label{fig:connections}
\end{figure}

For simplicity, we let $\log(\bullet)$ denote $\log_2(\bullet)$. 
In a proposed \logdensename, each layer $i$ takes direct input from at most $\log(i)+1$ number of previous layers, and these input layers are exponentially apart from depth $i$ with base $2$, i.e.,
\begin{align}
    \label{eq:logdense_layer}
    x_i = f_i(\concat(
    \{ x_{i - \lfloor 2^k \rceil} : k = 0, ..., \lfloor \log( i) \rfloor  \}
    )\;; \;\theta_i),
\end{align}
where $\lfloor \bullet \rceil$ is the nearest integer function and $\lfloor \bullet \rfloor$ is the floor function. For example, the input features for layer $i$ are layer $i-1, i-2, i-4,...$. We define the input index set at layer $i$ to be 
\mbox{$ \{ {i - \lfloor 2^k \rceil} : k = 0, ..., \lfloor \log( i) \rfloor  \}$}. We illustrate the connection in Fig.~\ref{fig:logv1_connections}. 
Since the complexity of layer $i$ is $\log(i)+1$, the overall complexity of a \logdense is $\sum _{i=1}^L (\log(i) + 1)\leq L + L\log L = \Theta( L \log L)$, which is significantly smaller than the quadratic complexity, $\Theta(L^2)$, of a DenseNet.


\textbf{\logdense V1: independent transition.}
Following \cite{densenet}, we organize layers into blocks. Layers in the same block have the same resolution; the feature map side is halved after each block. In between two consecutive blocks, a transition layer will shrink all previous layers so that future layers can use them in Eq~\ref{eq:logdense_layer}. We define a \textit{pooling transition} as a 1x1 conv followed by a 2x2 average pooling, where the output channel size of the conv is the same as the input one. 
We refer to $x_i$ after $t$ number of pooling transition as $x_i^{(t)}$. In particular, $x_i^{(0)} = x_i$. Then at each transition layer, for each $x_i$, we find the latest $x_i^{(t)}$, i.e., $t = max \{ s \geq 0 : x_i^{(s)} \text{exists}\}$, and compute $x_i^{(t+1)}$. 
We abuse the notation $x_i$ when it is used as an input of a feature layer to mean the appropriate $x_i^{(t)}$ so that the output and input resolutions match.
Unlike DenseNet, we independently process each early layer instead of using a pooling transition on the concatenated early features, because the latter option results in $O(L^2)$ complexity per transition layer, if at least $O(L)$ layers are to be processed. Since \logdense costs $O(L)$ computation for each transition, the total transition cost is $O(L\log L)$ as long as we have $O(\log L)$ transitions.

\textbf{\logdense V2: block compression.}
Unfortunately, many neural network packages, such as TensorFlow, cannot compute the $O(L)$ 1x1 conv for transition efficiently: in practice, this $O(L)$ operation costs about the same wall-clock time as the $O(L^2)$-cost 1x1 conv on the concatenation of the $O(L)$ layers.
To speed up transition and to further reduce \pbd, we propose a block compression for \logdense similar to the block compression in DenseNet~\citep{densenet}. At each transition, the newly finished block of feature layers are concatenated and compressed into $g \log L$ channels using 1x1 conv. The other previous compressed features are concatenated, followed by a 1x1 conv that keep the number of channels unchanged. These two blocks of compressed features then go through 2x2 average pooling to downsample, and are then concatenated together. Fig.~\ref{fig:logv2_connections} illustrates how the compressed features are used when $n_{block}=3$, where $x_0$, the initial conv layer of channel size $2g$, is considered the initial compressed block. The total connections and run-time complexity are still $O(L\log L)$, at any depth the total channel from the compressed feature is at most $(n_{block} -1) g \log L + 2g$, and we assume $n_{block} \leq 4$ is a constant. Furthermore, these transitions cost $O(L\log L)$ connections and computation in total, since compressing of the latest block costs $O(L\log L)$ and transforming the older blocks costs $O(\log^2 L)$.

\textbf{\logdense \pbd.} 
The reduction in complexity from $O(L^2)$ in DenseNet to $O(L\log L)$ in \logdense only increases the \pbd among layers to $1+\log L$. This result is summarized as follows.

\begin{proposition}
\label{them:log-dense-log-dist}
For any two feature layers $x_i \neq x_j$ in \logdense that has $n_{block}$ number of blocks, the \pbdfull between $x_i$ and $x_j$ is at most \mbox{ $ \log |j-i| + n_{block}$ }.
\end{proposition}

This proposition argues that if we ignore pooling layers, or in the case of \logdense V1, consider the transition layers as part of each feature layer, then any two layers $x_i$ and $x_j$ are only $\log |j-i| +1$ away from each other during backpropagation, so that layers can still easily affect each other to fit the training signals. 
Sec.~\ref{sec:exp-low-backprop-distance} experimentally shows that with the same amount the connections, the connection strategy with smaller \pbd leads to better accuracy. We defer the proof to the appendix. In comparison to \logdense V1, V2 reduces the \bd between any two layers from different blocks to be at most $n_{block}$, where the shortest paths go through the  compressed blocks.

It is also possible to provably achieve $2+\log \log L$ \pbd using only $1.5L\log\log L + o(L\log\log L)$ shortcut connections (Fig.~\ref{fig:loglog_connections}), but we defer this design to the appendix, because it has a complex construction and involves other factors that affect the prediction accuracy.

\textbf{Deep supervision.}
Since we cut the majority of the connections in DenseNet when forming \logdense, we found that having additional training signals at the intermediate layers using deep supervision \citep{supervisednet}  for the early layers helps the convergence of the network, even though the original DenseNet does not see performance impact from deep supervision. For simplicity, we place the auxiliary predictions at the end of each block. Let $x_i$ be a feature layer at the end of a block. Then the auxiliary prediction at $x_i$ takes as input $x_i$ along with $x_i$'s input features. Following \citep{hu:ann}, we put half of the total weighting in the final prediction and spread the other half evenly. After convergence, we take one extra epoch of training optimizing only the final prediction. We found this results in the lower validation error rate than always optimizing the final loss alone.

\section{Experiments}
\label{sec:exp-logense-main}

For visual recognition, we experiment on CIFAR10, CIFAR100 \citep{cifar}, SVHN \citep{svhn}, and ILSVRC2012~\citep{ilsvrc}.\footnote{CIFAR10 and CIFAR100 have 10 and 100 classes, and each have 50,000 training and 10,000 testing 32x32 color images. We adopt the standard augmentation to randomly flip left to right and crop 28x28 for training. SVHN contains around 600,000 training and around 26,000 testing 32x32 color images of numeric digits from the Google Street Views. We adopt the same pad-and-crop augmentations, and also apply Gaussian blurs. ILSVRC consists of 1.2 million training and 50,000 validation images from 1000 classes. We apply the same data augmentation for training as \citep{resnet, densenet}, and we report validation-set error rate from a single-crop of size 224x224 at test time. }
We follow \citep{resnet, densenet} for the training procedure and parameter choices. Specifically, we optimize using stochastic gradient descent with a moment of 0.9 and a batch size of 64 on CIFAR and SVHN. The learning rate starts at 0.1 and is divided by 10 after 1/2 and 3/4 of the total iterations are done. We train 250 epochs on CIFAR, 60 on SVHN, and 90 on ILSVRC. For CIFAR and SVHN, we specify a network by a pair $(n,g)$, where $n$ is the number of dense layers in each of the three dense blocks, and $g$, the growth rate, is the number of channels in each new layer.


\subsection{It matters where shortcut connections are}

\begin{table}
    \centering
    \resizebox{0.75\textwidth}{!}{
    \begin{tabular}{c|ccc| ccc| ccc}
   & \multicolumn{3}{c|}{CIFAR10} & \multicolumn{3}{c|}{CIFAR100}  & \multicolumn{3}{c}{SVHN}\\
\hline
 (n,g)   & L & N & E & L & N & E & L & N & E \\
\hline
(12,16)
	&  \textbf{7.23} &  7.59 &  7.45 & \textbf{29.14} & 30.59 & 30.72 &  \textbf{2.03} &  2.11 &  2.27 \\
(12,24)
	&  \textbf{5.98} &  6.46 &  6.56 & \textbf{26.36} & 26.96 & 27.80 &  \textbf{1.94} &  2.10 &  2.05 \\
(12,32)
	&  \textbf{5.48} &  6.00 &  6.15 & \textbf{24.21} & 24.70 & 25.57 &  \textbf{1.85} &  1.92 &  1.90 \\
\hline
(32,16)
	&  \textbf{5.96} &  6.45 &  6.21 & \textbf{25.32} & 27.48 & 26.81 &  1.97 &  \textbf{1.94} &  1.96 \\
(32,24)
	&  \textbf{5.03} &  5.74 &  5.43 & \textbf{22.73} & 25.08 & 24.80 &  \textbf{1.77} &  1.82 &  1.95 \\
(32,32)
	&  \textbf{4.81} &  5.65 &  4.94 & \textbf{21.77} & 23.79 & 23.87 &  \textbf{1.76} &  1.82 &  1.95 \\
\hline
(52,16)
	&  \textbf{5.13} &  6.80 &  6.09 & \textbf{23.45} & 27.99 & 26.58 &  \textbf{1.66} &  1.98 &  1.85 \\
(52,24)
	&  \textbf{4.34} &  5.83 &  5.03 & \textbf{20.99} & 26.07 & 24.19 &  \textbf{1.64} &  1.90 &  1.80 \\
(52,32)
	&  \textbf{4.56} &  6.10 &  4.98 & \textbf{20.58} & 24.79 & 23.10 &  \textbf{1.72} &  1.89 &  1.78 \\
\hline
    \end{tabular}}
    \caption{Error rates of \logdense V1(L), \nearest(N) and \evenspace(E), in each of which layer $x_i$ has $\log i$ 
    previous layers as input. (L) has a \pbd of $1+\log L$, and the other two have $\frac{L}{\log L}$.
    (L) outperforms the other two clearly. These networks do not have bottlenecks. }
    \label{tab:dsm_maintext}
\vspace{-8pt}
\end{table}

\begin{table}
\centering

    \subfloat[Error Rates with $i/2$ shotcuts to $x_i$]{
    \resizebox{0.6\textwidth}{!}{
    \begin{tabular}{c|ccc | ccc}
& \multicolumn{3}{c|}{CIFAR10} & \multicolumn{3}{c}{CIFAR100}  \\
\hline
 (n,g)   & N & E & N+L & N & E & N+L \\
\hline
(12,16)
	&  9.45 &  6.42 &  \textbf{5.77} & 35.97 & 29.65 & \textbf{25.49} \\
(12,24)
	&  6.49 &  5.18 &  \textbf{5.12} & 29.11 & 24.61 & \textbf{22.87} \\
(12,32)
	&  5.01 &  4.84 &  \textbf{4.70} & 25.04 & 23.70 & \textbf{21.96} \\
\hline
(32,12)
	&  7.16 &  4.90 &  \textbf{4.80} & 33.64 & 24.03 & \textbf{22.70} \\
(32,24)
	&  4.69 &  \textbf{4.16} &  4.36 & 24.58 & \textbf{21.00} & 21.27   \\
(32,32)
	&  4.30 &  4.24 &  \textbf{4.03} & 22.84 & \textbf{21.28} & 21.72  \\
\hline
(52,16)
	&  5.68 &  4.72 &  \textbf{4.34} & 28.44 & 21.73 & \textbf{20.68} \\
\hline 
    \end{tabular}}}
     ~
    \subfloat[NearestHalfAndLog]{\raisebox{-47pt}{
        \includegraphics[width=0.22\textwidth]{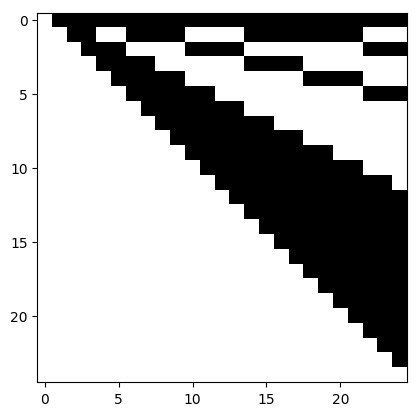}}
        \label{fig:nearest_half_and_log}
    }
    \caption{\textbf{(a)} \nearest (N), \evenspace (E), and NearestHalfAndLog (N+L) each connects to about $i/2$ previous layers at $x_i$, and have \pbd $\log L$, $2$ and $2$. N+L and E clearly outperform N. 
    \textbf{(b)} Connection illustration of N+L: each layer $i$ receives $\frac{i}{2} + \log(i)$ shortcut connections.}
    \label{tab:dsm678_maintext}

\vspace{-8pt}
\end{table}

\label{sec:exp-low-backprop-distance}
This section verifies that short \pbd is an important design principle by comparing the proposed \logdense V1 against two other intuitive connection strategies that also connects each layer $i$ to  $1 + \log(i)$ previous layers. The first strategy, called \textbf{\nearest} connects layer $i$ to its previous $\log(i)$ depths, i.e., 
\mbox{$x_i = f_i(\concat(
    \{ x_{i - k} : k = 1, ..., \lfloor \log_b( i) \rfloor  \}
    )\;; \;\theta_i).
$}
The second strategy, called \textbf{\evenspace} connects layer $i$ to $log(i)$ previous depths that are evenly spaced; i.e., 
\mbox{$x_i = f_i(\concat(
    \{ x_{ \lfloor i - 1 - k \delta \rceil} : 
    \delta = \frac{i}{log(i)} \text{ and }
    k =0,1,2,... \text{ and } k\delta \leq i-1 \}
    )\;; \;\theta_i).
$}
Both methods above are intuitive. However, each of them has a \pbd that is on the order of $O(\frac{L}{log(L)})$, which is much higher than the $O(log(L))$ \pbd of the proposed \logdense V1. 
We experiment with networks whose $(n,g)$ are in $\{12, 32, 52\} \times \{16, 24, 32\}$, and show in Table~\ref{tab:dsm_maintext} that \logdense almost always outperforms the other two strategies. Furthermore, the average relative increase of top-1 error rate using \nearest and \evenspace from using \logdense is $12.2\%$ and $8.5\%$, which is significant: for instance, (52,32) achieves 23.10\% error rate using \evenspace, which is about 10\% relatively worse than the 20.58\% from (52,32) using \logdense, but (52,16) using \logdense already has 23.45\% error rate using a quarter of the computation of (52,32). 

We also showcase the advantage of small \pbd when each layer $x_i$ is connects to $\approx \frac{i}{2}$ number of previous layers. With this many connections, \nearest has a \pbd of $\log L$, because we can halve $i$ (assuming $i > j$) until $j > i /2$ so that $i$ and $j$ are directly connected. \evenspace has a \pbd of $2$, because each $x_i$ takes input from every other previous layer.  Table~\ref{tab:dsm678_maintext} shows that \evenspace significantly outperform \nearest on CIFAR10 and CIFAR100.  We also show that \nearest is not under-performing simply because connecting to the most recent layers are ineffective. Starting with the \nearest scheme, we make $x_i$ also take input from $x_{\round{i/4}}, x_{\round{i/8}}, x_{\round{i/16}},...$. We call this scheme NearestHalfAndLog, and it has a \pbd of $2$, because any $j < i$ is either directly connected to $i$, if $j > i / 2$, or $j$ is connected to some $i / \round{i / 2^k}$ for some $k$, which is connected to $i$ directly. 
Fig.~\ref{fig:nearest_half_and_log} illustrates the connections of this scheme.
We observe in Table~\ref{tab:dsm678_maintext} that with this few $\log_i -1$ additional connections to the existing $\lceil i / 2 \rceil$ ones, we drastically reduce the error rates to the level of \evenspace, which has the same \pbd of 2. 
These comparisons support our design principle: with the same number of connections at each depth $i$, the connection strategies with low \pbd outperform the ones with high \pbd.

\subsection{\logdense for Training Semantic Segmentation from Scratch}
\label{sec:exp-fcdense}
\begin{table}[t]
\resizebox{\textwidth}{!}{
\begin{tabular}{c| c |c | *{11}{|c} || c| c}
\centering

\rotatebox{90}{ Method } & \rotatebox{90}{ GFLOPS } & \rotatebox{90}{ \# Params (M) } & \rotatebox{90}{ Building } & \rotatebox{90}{ Tree } & \rotatebox{90}{ Sky } & \rotatebox{90}{ Car } & \rotatebox{90}{ Sign } & \rotatebox{90}{ Road } & \rotatebox{90}{ Pedestrian } & \rotatebox{90}{ Fence } & \rotatebox{90}{ Pole } & \rotatebox{90}{ Sidewalk } & \rotatebox{90}{ Cyclist } & \rotatebox{90}{ Mean IoU } & \rotatebox{90}{ Accuracy } \\
\hline
SegNet$^1$
 & 	- &  29.5
 &  68.7 &  52.0 &  87.0 &  58.5 &  13.4 &  86.2 &  25.3 &  17.9 &  16.0 &  60.5 &  24.8 &  46.4 &  62.5 \\
\hline
FCN8$^2$ 
 & - & 134.5
 &  77.8 &  71.0 &  88.7 &  76.1 &  32.7 &  91.2 &  41.7 &  24.4 &  19.9 &  72.7 &  31.0 &  57.0 &  88.0 \\
\hline
DeepLab-LFOV$^3$ 
 & - &  37.3
 &  81.5 &  74.6 &  89.0 &  82.2 &  42.3 &  92.2 &  48.4 &  27.2 &  14.3 &  75.4 &  50.1 &  61.6 &   nan \\
\hline 
Dilation8 + FSO$^4$ 
 & - & 140.8
 & \textbf{  84.0 } & \textbf{  77.2 } &  91.3 & \textbf{  85.6 } & \textbf{  49.9 } &  92.5 &  59.1 &  37.6 &  16.9 &  76.0 & \textbf{  57.2 } &  66.1 &  88.3 \\
\hline 
FC-DenseNet67 (g=16)$^5$ 
 &  40.9 &   3.5
 &  80.2 &  75.4 & \textbf{  93.0 } &  78.2 &  40.9 & \textbf{  94.7 } &  58.4 &  30.7 & \textbf{  38.4 } &  81.9 &  52.1 &  65.8 &  90.8 \\
\hline 
FC-DenseNet103 (g=16)$^5$
 & 39.4 &   9.4
 &  83.0 & \textbf{  77.3 } & \textbf{  93.0 } &  77.3 &  43.9 &  94.5 & \textbf{  59.6 } &  37.1 &  37.8 & \textbf{  82.2 } &  50.5 &  66.9 &  91.5 \\
\hline 
LogDensenetV1-103 (g=24)
 &  42.0 &   4.7
 &  81.6 &  75.5 &  92.3 &  81.9 &  44.4 &  92.6 &  58.3 & \textbf{  42.3 } &  37.2 &  77.5 &  56.6 & \textbf{  67.3 } &  90.7 \\
\hline
\end{tabular}}
\caption{Performance on the CamVid semantic segmentation data-set. The column GFLOPS reports the computation on a 224x224 image in 1e9 FLOPS. We compare against 1~\citep{segnet}, 2~\citep{fcn}, 3~\citep{deeplab}, 4~\citep{fso}, and 5~\citep{fcdense}.}
\label{tab:camvid_ious}
\vspace{-10pt}
\end{table}

Semantic segmentation assigns every pixel of input images with a label class, and it is an important step for understanding image scenes for robotics such as autonomous driving. The state-of-the-art training procedure~\citep{pspnet, deeplab} typically requires training a fully-convolutional network (FCN)~\citep{fcn} and starting with a recognition network that is trained on large data-sets such as ILSVRC or COCO, because training FCNs from scratch is prone to overfitting and is difficult to converge.
\cite{fcdense} shows that DenseNets are promising for enabling FCNs to be trained from scratch. In fact, fully convolutional DenseNets (FC-DenseNets) are shown to be able to achieve the state-of-the-art predictions training from scratch without additional data on CamVid \citep{camvid-dataset} and GATech \citep{gatech-dataset}. However, the drawbacks of DenseNet are already manifested in applications on even relatively small images (360x480 resolution from CamVid). In particular, to fit FC-DenseNet into memory and to run it in reasonable speed, \cite{fcdense} proposes to cut many mid-connections: during upsampling, each layer is only directly connected to layers in its current block and its immediately previous block. Such connection strategy is similar to the \nearest strategy in Sec.~\ref{sec:exp-low-backprop-distance}, which has already been shown to be less effective than the proposed \logdense in classification tasks. We now experimentally show that fully-convolutional \logdense (FC-\logdense) outperforms FC-DenseNet.

\textbf{FC-\logdense 103.} Following \citep{fcdense}, we form FC-\logdense V1-103 with 11 \logdense V1 blocks, where the number of feature layers in the blocks are $4, 5, 7, 10, 12, 15, 12, 10, 7, 5, 4$. After each of the first five blocks, there is a transition that transforms and downsamples previous layers independently. After each of the next five blocks, there is a transition that applies a transposed convolution to upsample each previous layer. Both down and up sampling are only done when needed, so that if a layer is not used directly in the future, no transition is applied to it. Each feature layer takes input using the \logdense connection strategy. Since \logdense connections are sparse to early layers, which contain important high resolution features for high resolution semantic segmentation, we add feature layer $x_4$, which is the last layer of the first block, to the input set of all subsequent layers. This adds only one extra connection for each layer after the first block, so the overall complexity remains roughly the same. 
We do not form any other skip connections, since \logdense  already provides sparse connections to past layers. 
 
\begin{figure}
    \centering
    \subfloat{
    \includegraphics[width=\linewidth]{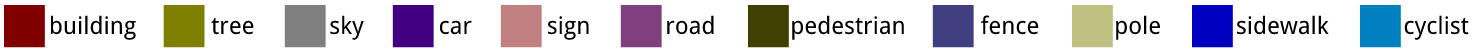}
    }
    
    \subfloat{
    \includegraphics[width=\linewidth]{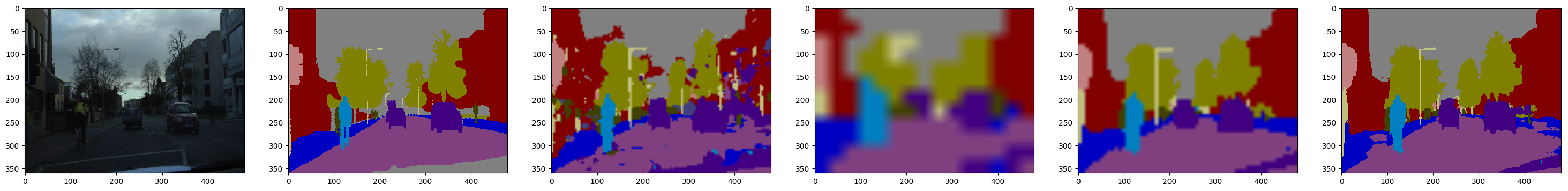}
    }
    
    \subfloat{
    \includegraphics[width=\linewidth]{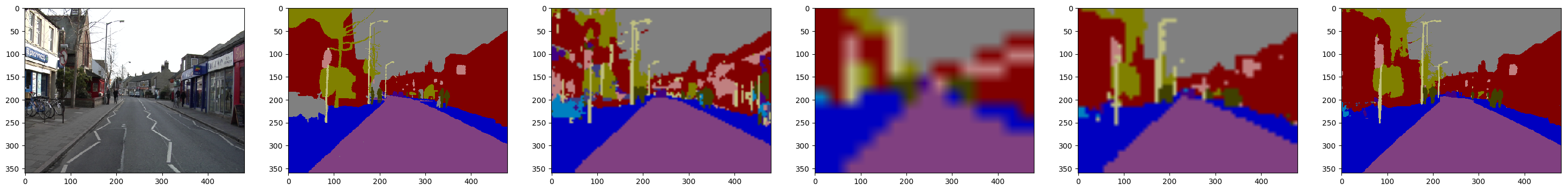}
    }

    \caption{Each row: input image, ground truth labeling, and any scene parsing results at 1/4, 1/2, 3/4 and the final layer. The prediction at 1/2 is blurred, because it and its feature are at a low resolution.}
    \label{fig:scene_parsing}
    \vspace{-5pt}
\end{figure}

\textbf{Training details.} Our training procedure and parameters follow  from those of FC-DenseNet~\citep{fcdense}, except that we set the growth rate to 24 instead of 16, in order to have around the same computational cost as FC-DenseNet. 
We defer the details to the appendix. However, we also found auxiliary predictions at the end of each dense block reduce overfitting and produce interesting progression of the predictions, as shown in Fig.~\ref{fig:scene_parsing}. Specifically, these auxiliary predictions produces semantic segmentation at the scale of their features using 1x1 conv layers. The inputs of the predictions and the weighting of the losses are the same as in classification, as specified in Sec.~\ref{sec:logdense}.

\textbf{Performance analysis.}
We note that the final two blocks of FC-DenseNet and FC-\logdense cost half of their total computation. This is because the final blocks have fine resolutions, which also make the full DenseNet connection in the final two blocks prohibitively expensive. This is also why FC-DenseNets~\citep{fcdense} have to forgo all the mid-depth the shortcut connections in its upsampling blocks. 
Table~\ref{tab:camvid_ious} lists the Intersection-over-Union ratios (IoUs) of the scene parsing results.
FC-\logdense achieves 67.3\% mean IoUs, which is slightly higher than the 66.9\% of FC-DenseNet. Among the 11 classes, FC-\logdense performs similarly to FC-DenseNet. Hence FC-\logdense achieves the same level of performance as FC-DenseNet with 50\% fewer parameters and similar computations in FLOPS. This supports our hypothesis that we should minimize \pbd when we have can only have a limited number of skip connections. FC-\logdense can potentially be improved if we reuse the shortcut connections in the final block to reduce the number of upsamplings.

\subsection{Computational Efficiency of Sparse and Dense Networks}
\label{sec:exp-trade-off}
\vspace{-5pt}

\begin{figure}[t]
    \centering
    \subfloat[CIFAR100 Error versus FLOPS]{
        \includegraphics[width=0.45\textwidth]{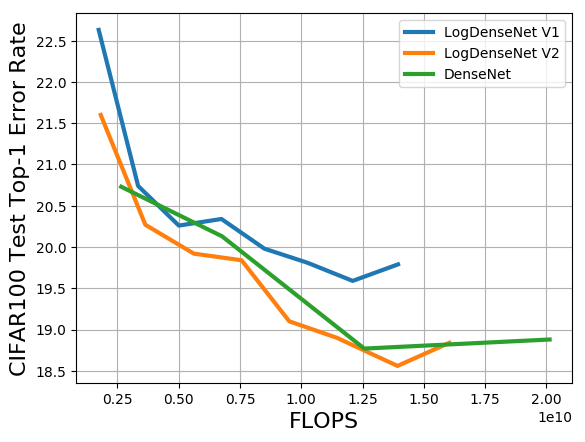}
        \label{fig:flops_cifar100}
    }
    ~
    \subfloat[ILSVRC Error versus FLOPS]{
        \includegraphics[width=0.45\textwidth]{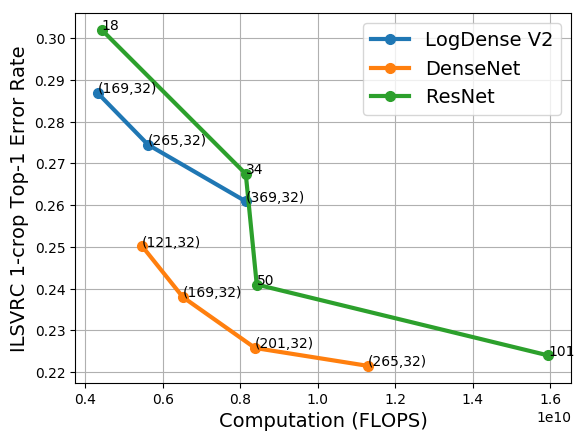}
        \label{fig:flops_ilsvrc}
    }
    \caption{\textbf{(a)} Using the same FLOPS, \logdense V2 achieves about the same prediction accuracy as DenseNets on CIFAR100. The DenseNets have block compression and are trained with drop-outs. \textbf{(b)} On ILSVRC2012, \logdense169, 265 have the same block sizes as DenseNets169, 265. \logdensenp369 has block sizes $8,16, 80, 80$. }
    \vspace{-10pt}
\end{figure}

This section studies the trade-off between computational cost and the accuracy of networks on visual recognition. In particular, we address the question of whether sparser networks like \logdense perform better than DenseNet using the same computation. DenseNets can be very deep for image classification, because they have low resolution in the final block. In particular, a skip connection to the final block costs $1/64$ of one to the first block. 
Fig.~\ref{fig:flops_cifar100} illustrates the error rates on CIFAR100 of \logdense V1 and V2 and DenseNet. The \logdense variants have $g=32$, and $n=12,22,32,...,82$. DenseNets have $g=32$, and $n=12,22,32,42$. \logdense V2 has around the same performance as DenseNet on CIFAR100. This is partially explained by the fact that most pairs of $x_i, x_j$ in \logdense V2 are cross-block, so that they have the same \pbd as in Densenets thanks to the compressed early blocks. The within block distance is bounded by the logarithm of the block size, which is smaller than 7 here.
\logdense V1 has similar error rates as the other two, but is slightly worse, an expected result, because unlike V2, backpropagation distances between a pair $x_i,x_j$ in V1 is always $ \log |i -j|$, so on average V1 has a higher \pbd than V2 does. 
The performance gap between \logdense V1 and DenseNet also gradually widens with the depth of the network, possibly because the \pbd of \logdense has a logarithmic growth. We observe similar effects on CIFAR10 and SVHN, whose performance versus computational cost plots are deferred to the appendix.
These comparisons suggest that to reach the same accuracy, the sparse \logdense costs about the same computation as the DenseNet, but is capable of scaling to much higher depths. 
We also note that using \naive implementations, and a fixed batch size of 16 per GPU, DenseNets (52, 24) already have difficulties fitting in the 11GB RAM, but \logdense can fit models with $n>100$ with the same $g$. We defer the plots for number of parameters versus error rates to the appendix as they look almost the same as plots for FLOPS versus error rates.

On the more challenging ILSVRC2012~\citep{ilsvrc}, we observe that \logdense V2 can achieve comparable error rates to DenseNet. Specifically, \logdense V2 is more computationally efficient than ResNet~\citep{resnet} that do not use bottlenecks (ResNet18 and ResNet34): \logdense V2 can achieve lower prediction errors with the same computational cost. However, \logdense V2 is not as computationally efficient as ResNet with bottlenecks (ResNet 50 and ResNet101), or DenseNet. This implies there may be a trade-off between the shortcut connection density and the computation efficiency. For problems where shallow networks with dense connections can learn good predictors, there may be no need to scale to very deep networks with sparse connections. However, the proposed \logdense  provides a reasonable trade-off between  accuracy and scalability for tasks that require deep networks, as in Sec.~\ref{sec:exp-fcdense}. 




\vspace{-5pt}
\section{Conclusions and Discussions}
\vspace{-5pt}
\label{sec:conclusion}
We show that short backpropagation distances are important for networks that have shortcut connections: if each layer has a fixed number of shortcut inputs, they should be placed to minimize \pbd. Based on this principle, we design \logdense, which uses $O(L \log L)$ total shortcut connections on a depth-$L$ network to achieve $1 + \log L$ \pbd. We show that \logdenses improve the performance and scalability of \textit{tabula rasa} fully convolutional DenseNets on CamVid. \logdenses also achieve competitive results in visual recognition data-sets, offering a trade-off between accuracy and network depth. Our work provides insights for future network designs, especially those that cannot afford full dense shortcut connections and need high depths, like FCNs.

\bibliography{ann}
\bibliographystyle{iclr2018_conference}

\newpage
\appendix

\textbf{Appendix}

\section{\loglogdense}
\label{sec:loglogdense}

\begin{figure}
    \centering
    \subfloat[Illustration of lglg\_conn recursion]{
        \includegraphics[width=0.45\textwidth]{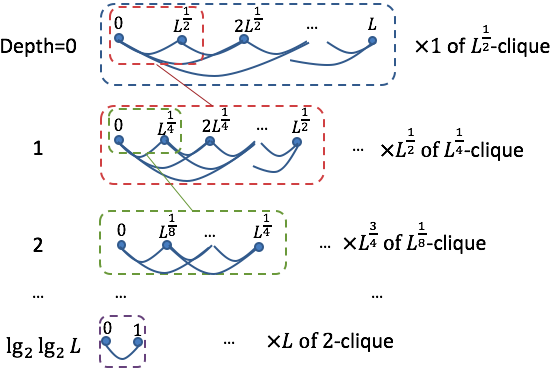}
    }
    ~
    \subfloat[lglg\_conn($0,L$)]{
        \includegraphics[width=0.22\textwidth]{LogLog_connections.png}
    }
    ~
    \subfloat[\loglogdense]{
        \includegraphics[width=0.22\textwidth]{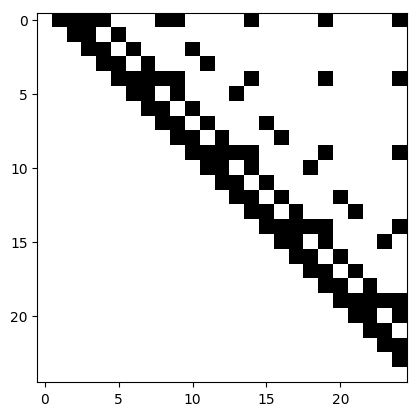}
    }
    \cprotect\caption{\textbf{(a)}The tree of recursive calls of \verb=lglg_conn=. \textbf{(b)} \loglogdense augments each $x_i$ of \verb=lglg_conn=($0,L$) with \logdense connections until $x_i$ has at least four inputs.
    }
    \label{fig:loglog_recursion}
\end{figure}
Following the short \pbd design principle, we further propose \loglogdense, which uses $O(L\log \log L)$ connections to achieve $1+\log\log L$ \pbd. 
For the clarity of construction, we assume there is a single block for now, i.e., $n_{block}=1$.
We add connections in \loglogdense recursively, after we initialize each depth $i=1,...,L$ to take input from $i-1$, where layer $i=0$ is the initial convolution that transform the image to a $2g$-channel feature map. Fig.~\ref{fig:loglog_recursion} illustrates the recursive calls. Formally, the recursive connection-adding function is called \verb=lglg_conn=$(s,t)$, where the inputs $s$ and $t$ represent the start and the end indices of a segment of contiguous layers in $0,...,L$. For instance, the root of the recursive call is \verb=lglg_conn=($0,L$), where $(0,L)$ represents the segment of all the layers $0,...,L$.
\verb=lglg_conn=$(s,t)$ exits immediately if $t-s \leq 1$. 
If otherwise, we let $\delta = \lfloor \sqrt{t-s+1} \rfloor$, $K=\{s \} \cup \{ t- k \delta : t - k \delta  \geq s \text{ and }  k =0,1,2,..., \}$, and let $a_1,...,a_{|K|}$ be the sorted elements of $K$.
\textbf{(a)} Then we add dense connections among layers whose indices are in $K$, i.e., if $i, j \in K$ and $i > j$, then we add $x_j$ to the input set of $x_i$. \textbf{(b)} Next for each $k=1,..., |K|-1$, we add $a_{k}$ to the input set of $x_j$ for each $j=a_{k}+1,..., a_{k+1}$. \textbf{(c)} Finally, we form $|K|-1$ number of recursive \verb=lglg_conn= calls, whose inputs are
$(s_k, s_{k+1})$ for each $k=1,...,|K|-1$. 

If $n_{block} > 1$, we reuse \logdense V1 transition to scale each layer independently, so that when we add $x_j$ to the input set of $x_i$, the appropriately scaled version of $x_j$ is used instead of the original $x_j$. 
We defer to the appendix the formal analysis on the recursion tree rooting at \verb=lglg_conn=($0,L$), which forms the connection in \loglogdense, and summarize the result as follows.
\begin{proposition}
\label{them:loglog-dense-loglog-dist}
\loglogdense of $L$ feature layers has at most $1.5L \log\log L + o(L \log\log L)$ connections, and a \pbd at most \mbox{ $ \log\log L + n_{block}$ +1 }.
\end{proposition}

Hence, if we ignore the independent transitions and think them as part of each $x_i$ computation, the \pbd between any two layers $x_i, x_j$ in \loglogdense is at most $2 + \log\log L$, which effectively equals 5, because  $\log\log L < 3.5$ for $L < 2545$. Furthermore, such short \pbd is very cheap: on average, each layer takes input from 3 to 4 layers for $L < 1700$, which we verify in~\ref{sec:loglogdense_proof}. We also note that without step (b) in the \verb=lglg_conn=, the \pbd is $2+ 2\log \log L$ instead of $2 + \log \log L$.

\textbf{Bottlenecks.} 
In DenseNet, since each layer takes input from the concatenation of all previous layers, it is necessary to have bottleneck structures \citep{resnet, densenet}, which uses 1x1 conv to shrink the channel size to $4g$ first, before using 3x3 conv to generate the $g$ channel of features in each $x_i$. In \logdense and \loglogdense, however, the number of input layers is so small that bottlenecks actually increase the computation, e.g., most of \loglogdense layers do not even have $4g$ input channels. However, we found that bottlenecks are cost effective for increasing the network depth and accuracy. Hence, to reduce the variation of structures, we use bottlenecks and fix the bottleneck width to be $4g$. For \loglogdense, we also add \logdense connections from nearest to farthest for each $x_i$ until either $x_i$ has four inputs, or there are no available layers. For $L<1700$, this increases average input sizes only to $4.5 \sim 5$, which we detail in the next section.

\section{Formal analysis of \logdense and \loglogdense}

\subsection{Proof of Proposition~\ref{them:log-dense-log-dist}}
\label{sec:logdense_proof}
\begin{myproof}
We call \bd$(x_i, x_j)$ the back-propagation distance from $x_i$, $x_j$, which is the distance between the two 
nodes $x_i, x_j$ on the graph constructed for defining \pbd in Sec.~\ref{sec:pbd}.
The scaling transition happens only once for each scale during backpropagation; i.e., there 
are at most $n_{block}-1$ number of transitions between any two layers $x_i$, $x_j$.  

Since the transition between each two scales happens at most once, in between two layers $x_i$, $x_j$, we first consider $n_{block}=1$, and add $n_{block}-1$ to the final distance bound to account for multiple blocks. 

We now prove the proposition for $n_{block}=1$ by induction on $|i-j|$. 
Without loss of generality we assume $i>j$. 
The base case: for all $i> j$ such that $i = j+1$, we have \bd$(x_i, x_j)$ = 1.
Now we assume the induction hypothesis that for some $t \geq 0$ and $t\in \mathbb{N}$, such that 
for all $i >j$ and $i - j \leq 2^t$, we have \bd$(x_i, x_j) \leq t +1$.  
Then for any two layers $i >j$ such that $i -j \leq 2^{t+1}$, if 
$i - j \leq 2^t$, then by the induction hypothesis, \bd$(x_i, x_j) \leq t +1 < t+2$.
If $2^{t+1} \geq i - j > 2^t$, then 
we have $k := i - 2^t > j$, and $k= i - 2^t \leq 2^{t+1} -2^t + j = j+ 2^t$. So that 
$k - j \in (0, 2^t]$. Next by the induction hypothesis, 
\bd$(x_k, x_j) \leq t+1$. 
Furthermore, by the connections of \logdense, $x_i$ takes input directly from 
$x_{i-2^t}$, so that \bd($x_i$, $x_k$) = 1. Hence, by the triangle inequality of 
distances in graphs, we have \bd$(x_i, x_j) \leq$ \bd$(x_i, x_k)$ + \bd$(x_k, x_j) \leq 1 + (t+1)$. 
This proves the induction hypothesis for $t+1$, so that the proposition follows, i.e., 
for any $i \neq j$, \bd$(x_i, x_j) \leq \log |i-j| + 1$. 

\end{myproof}


\subsection{Proof of Proposition~\ref{them:loglog-dense-loglog-dist}}
\label{sec:loglogdense_proof}

\begin{myproof}
Since the transition between each two scales happens at most once, in between two layers $x_i$, $x_j$, we again first consider $n_{block}=1$, and add $n_{block}-1$ to the final distance bound to account for multiple blocks. 

(\textbf{Number of connections.})
We first analyze the recursion tree of \verb=lglg_conn=($0,L$). 
In each \verb=lglg_conn=$(s,t)$ call, let $n = t-s+1$ be the number of layers on the segment $(s,t)$. 
Then the interval of key locations $\delta = \lfloor \sqrt{t-s+1} \rfloor = \lfloor \sqrt{n} \rfloor$, and 
the key location set $K=\{s\} \cup \{ t-   k \delta  : t- k \delta  \geq s \text{ and }  k =0,1,2,..., \} $ has 
a cardinality of $|K| = 1 + \lceil \frac{n-1}{\lfloor \sqrt{n} \rfloor} \rceil \in (\sqrt{n},  2.5 + \sqrt{n})$. 
Hence the step (a) of \verb=lglg_conn=$(s,t)$ in Sec.~\ref{sec:loglogdense} adds $1+2+3+... + (|K|-1) = 0.5|K|(|K|-1)$.
Step (b) of \verb=lglg_conn=$(s,t)$ then creates $(n - |K|)$ new connections, since the ones among $x_s,..., x_t$ that are not given new connections are exactly $x_i$ in $K$. Hence, \verb=lglg_conn=$(s,t)$ using step (a),(b) increases the total connections by 
\begin{align}
   c(n) = n  + 0.5 |K|^2 - 1.5|K| < 1.5n + \sqrt{n} + 3.125.
\end{align}
Step (c) instantiate $|K|-1 \leq \sqrt{n} + 1.5$ calls of \verb=lglg_conn=, each of which has an input segment of length at most $\delta \leq \sqrt{n}  + 1$.
Hence, let C(n) be the number of connections made by the recursive call \verb=lglg_conn=$(s,t)$ for $n = t-s+1$, then we have the recursion
\begin{align}
    C(n) \leq (\sqrt{n} + 1.5) C(\sqrt{n}+1) + c(n).
\end{align}
Hence, the input segment length takes a square root in each depth until the base case at length 2. 
The depth of the recursion tree of \verb=lglg_conn=$(s,t)$ is then $1 + \log \log (t-s+1)$. Furthermore, the connections made on each depth $i$ of the tree is $1.5n + o(n)$, because at each depth $i=0,1...$, 
$c(n^{2^{-i}} + o(n^{2^{-i}})) = 1.5n^{2^{-i}} + o(n^{2^{-i}})$ connectons are made in each \verb=lglg_conn=, and the number of calls is 1 for $i=0$, and 
$\Pi _{j=0}^{i} (n^{2^{-j}}+1.5) = n^{1 - 2^{-i}} + o(n^{1 - 2^{-i}})$. Hence the total connections in \loglogdense is $C(L+1) = 1.5L\log\log L + o(L\log\log L)$. 

(\textbf{Back-propogation distance.})
First, each $x_i$ for $i \in [s,t]$ in the key location set $K$ of \verb=lglg_conn=($s,t$) is in the input set of $x_t$. 
Second, for every $x_i$, and for every call \verb=lglg_conn=($s,t$) in the recursion tree such that $s < i \leq t$ and $(s,t) \neq (0,L)$, we know step (b) adds $x_s$ to the input set of $x_i$. Hence, we can form a back-propagation path from any $x_i$ to $x_j$ $(i  > j)$ by first using a connection from step (b) to go to a key location of the \verb=lglg_conn=($s,t$) call such that $[s,t]$ is the smallest interval in the recursion tree such that $i,j \in [s,t]$. Then we can continue the path to $x_j$ by following the recursion calls whose input segments include $j$ until $j$ is in a key location set. The longest path is then the depth of the recursion tree plus one initial jump, i.e., $2 + \log\log L$.

\end{myproof}


\subsection{\loglogdense Layers on Average Has Five Connections in Practice}
\begin{figure}
    \centering
    \subfloat[Average number of inputs per Layer in \loglogdense]{
        \includegraphics[width=0.5\textwidth]{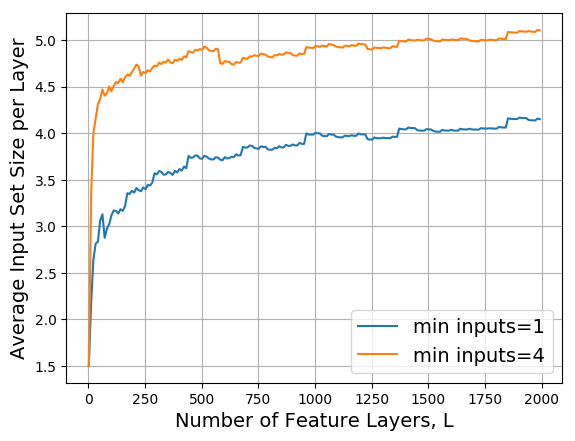}
        \label{fig:loglog_avg_connections}
    }
    ~
    \subfloat[Blocks versus Computational Cost in FCNs]{
        \includegraphics[width=0.5\textwidth]{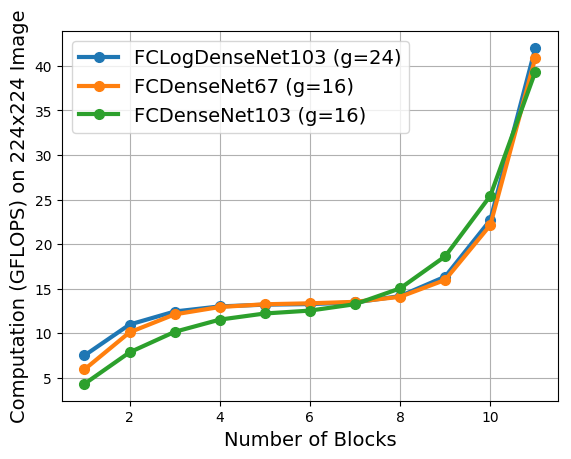}
        \label{fig:fc_computation}
    }

    \cprotect\caption{\textbf{(a)} In \verb=lglg_conn=$(0,L)$, i.e., min inputs = 1, each layer on average takes input from 3 to 4 layers. If we force input size to be four when possible using \logdense connection pattern, i.e., min inputs = 4, we increase the average input size by 1 to 1.5. \textbf{(b)} Computational cost (in FLOPS) distribution through the 11 blocks in FC-DenseNet and FC-\logdense. Half of the computations are from the final two blocks due to the high final resolutions. We compute the FLOPS assuming the input is a single 224x224 image.}
    \label{fig:my_label}
\end{figure}

Figure~\ref{fig:loglog_avg_connections} shows the average number of input layers for each feature layer in \loglogdense. Without augmentations, \verb=lglg_conn= on average has 3 to 4 connections per layer. With augmentations using \logdense, we desire each layer to have four inputs if possible. On average, this increases the number of inputs by 1 to 1.5 for $L \in (10, 2000)$.

\section{Additional Experimental Results}

\subsection{CamVid Training Details}
We follow \cite{fcdense} to optimize the network using 224x224 random cropped images with RMSprop. The learning rate is 1e-3 with a decay rate 0.995 for 700 epochs. We then fine-tune on full images with a learning rate of 5e-4 with the same decay for 300 epochs.  The batch size is set to 6 during training and 2 during fine-tuning. We train on two GTX 1080 GPUs. 
We use no pre-processing of the data, except left-right random flipping. Following \cite{segnet}, we use the median class weighting to balance the weights of classes, i.e., the weight of each class $C$ is the median of the class probabilities divided by the over the probability of $C$.

\subsection{Computational Efficiency on CIFAR10 and SVHN}

\begin{figure}[t]
    \centering
    \subfloat[CIFAR10 Error versus FLOPS]{
        \includegraphics[width=0.42\textwidth]{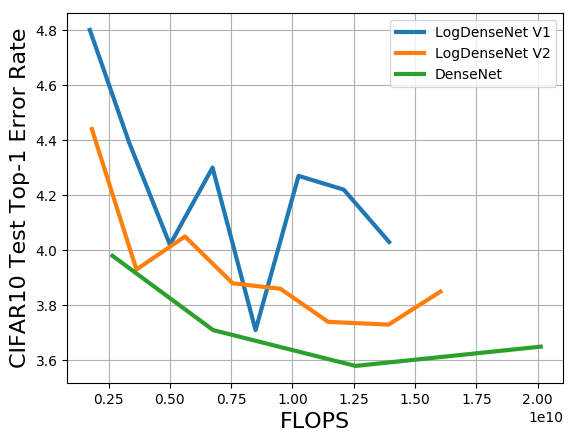}
        \label{fig:flops_cifar10}
    }
    ~
    \subfloat[SVHN Error versus FLOPS]{
        \includegraphics[width=0.42\textwidth]{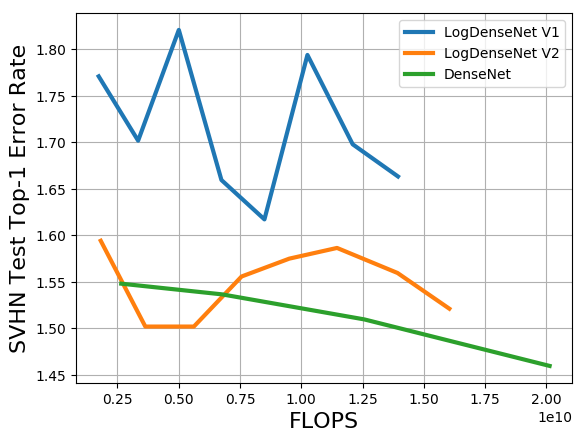}
        \label{fig:flops_svhn}
    }
    \caption{On CIFAR10 and SVHN, \logdense V2 and DenseNets have very close error rates ($<5\%$ relatively difference) at each budget.}
\end{figure}

Fig.~\ref{fig:flops_cifar10} and Fig.~\ref{fig:flops_svhn} illustrate the trade-off between computation and accuracy of \logdense and DenseNets on CIFAR10 and SVHN. \logdenses V2 and DenseNets have similar performances on these data-sets: on CIFAR10, the error rate difference at each budget is less than 0.2\% out of 3.6\% total error; on SVHN, the error rate difference is less than 0.05\% out of 1.5\%. Hence, in both cases, the error rates between \logdense V2 and DenseNets are around 5\%. 

\subsection{Number of Parameter  versus Error Rates. }
\begin{figure}
    \centering
    \subfloat[CIFAR10 Number of Parameters versus Error Rates]{
        \includegraphics[width=0.42\textwidth]{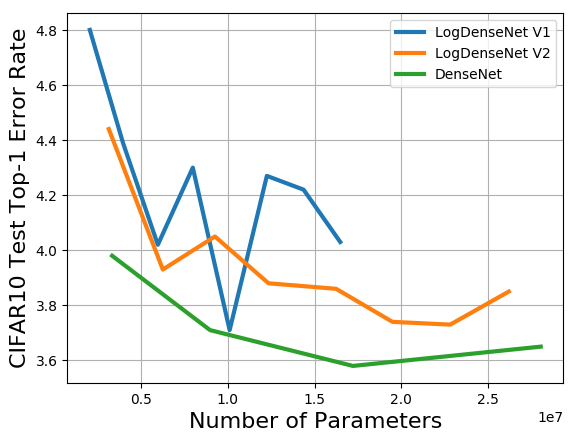}
        \label{fig:params_cifar10}
    }
    ~
    \subfloat[CIFAR100 Number of Parameters versus Error Rates]{
        \includegraphics[width=0.42\textwidth]{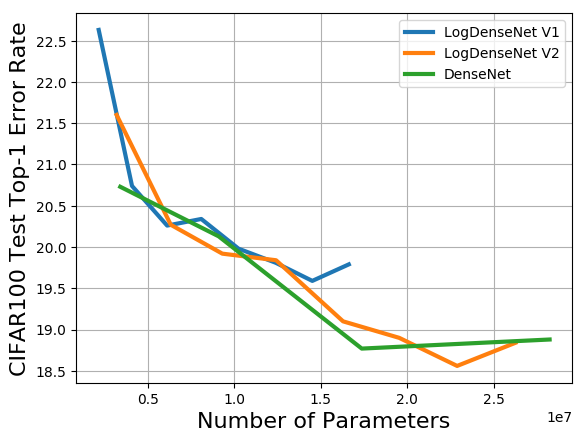}
        \label{fig:params_cifar100}
    }
    
    \subfloat[SVHN Number of Parameters versus Error Rates]{
        \includegraphics[width=0.42\textwidth]{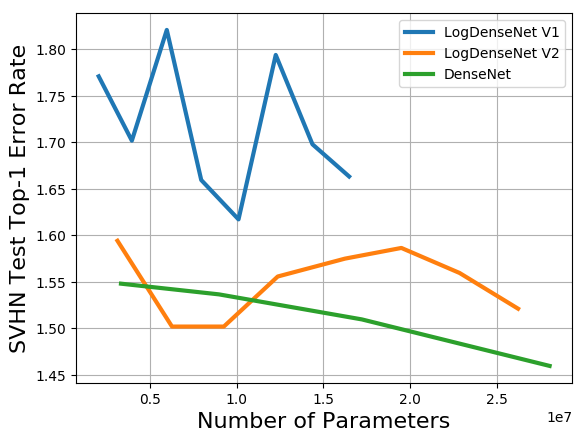}
        \label{fig:params_svhn}
    }
    ~
    \subfloat[ILSVRC Number of Parameters versus Error Rates]{
        \includegraphics[width=0.42\textwidth]{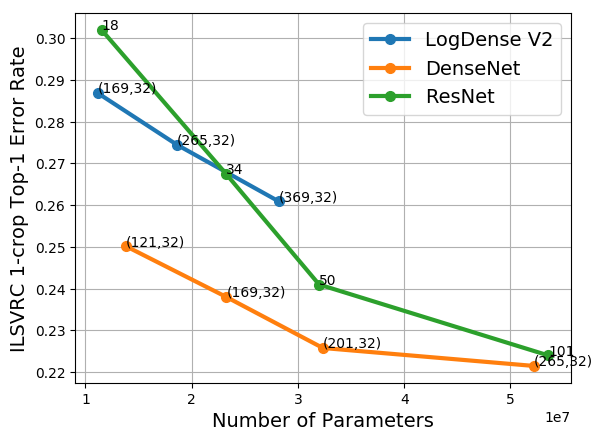}
        \label{fig:params_ilsvrc}
    }
    \caption{The number of parameter used in the \naive implementation versus the error rates on various data-sets.}
    \label{fig:params_all_data}
\end{figure}

Figure~\ref{fig:params_all_data} plots the number of parameters used by \logdense V2, DenseNet, and ResNet versus the error rates on the image classification data-sets, CIFAR10, CIFAR100, SVHN, ILSVRC. We assume that DenseNet and \logdense use \naive implementations. 

\subsection{\loglogdense Experiments and More Principles than \pbd}
\begin{figure}
    \centering
    \subfloat[\loglogdense Hub Multiplier=1]{
    \includegraphics[width=0.42\linewidth]{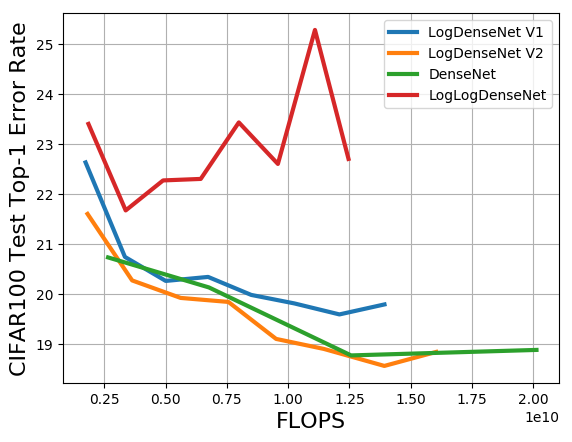}
    \label{fig:loglog_llm1}
    }
    ~
    \subfloat[\loglogdense Hub Multiplier=3]{
    \includegraphics[width=0.42\linewidth]{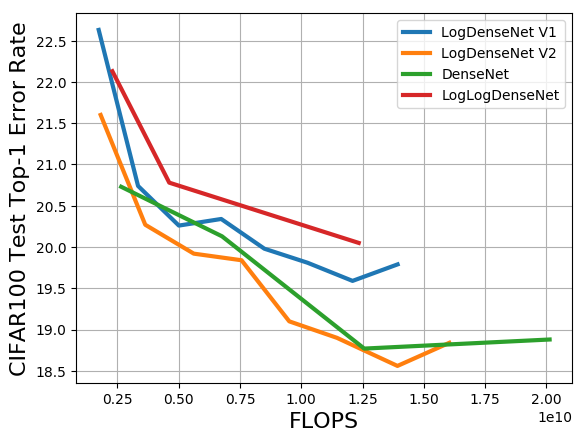}
    \label{fig:loglog_llm3}
    }

\caption{Performance of \loglogdense (red) with different hub multiplier (1 and 3). Larger hubs allow more information to be passed by the hub layers, so the predictions are more accurate.}
\end{figure}

This section experiments with \loglogdense and show that there are more that just \pbd that affects the performance of networks. Ideally, since \loglogdense have very small \pbd, its performance should be very close to DenseNet, if \pbd is the sole decider of the performance of networks. However, we observe in Fig.~\ref{fig:loglog_llm1} that \loglogdense is not only much worse than \logdense and DenseNet in terms accuracy at each given computational cost (in FLOPS), it is also widening the performance gap to the extent that the test error rate actually increases with the depth of the network. This suggests there are more factors at play than just \pbd, and in deep \loglogdense, these factors inhibit the networks from converging well. 

One key difference between \loglogdense's connection pattern to \logdense's is that the layers are not symmetric, in the sense that layers have drastically different shortcut connection inputs. In particular, while the average input connections per layer is five (as shown in Fig.~\ref{fig:loglog_avg_connections}), some nodes, such as the nodes that are multiples of $L^{\frac{1}{2}}$, have very large in-degrees and out-degrees (i.e., the number of input and output connections). These nodes are given the same number of channels as any other nodes, which means there must be some information loss passing through such ``hub" layers, which we define as layers that are densely connected on the depth zero of \verb=lglg_conn= call. 
Hence a natural remedy is to increase the channel size of the hub nodes. In fact, Fig.~\ref{fig:loglog_llm3} shows that by giving the hub layers three times as many channels, we greatly improve the performance of \loglogdense to the level of \logdense. This experiment also suggests that the layers in networks with shortcut connections should ensure that high degree layers have enough capacity (channels) to support the amount of information passing.

\subsection{Additional Semantic Segmentation Results}
\label{sec:additional_scene_parsing}
\begin{figure}
    \centering
    \subfloat{
    \includegraphics[width=\linewidth]{camvid_cmap.png}
    }
    
    \subfloat{
    \includegraphics[width=\linewidth]{img_0.png}
    }
    
    \subfloat{
    \includegraphics[width=\linewidth]{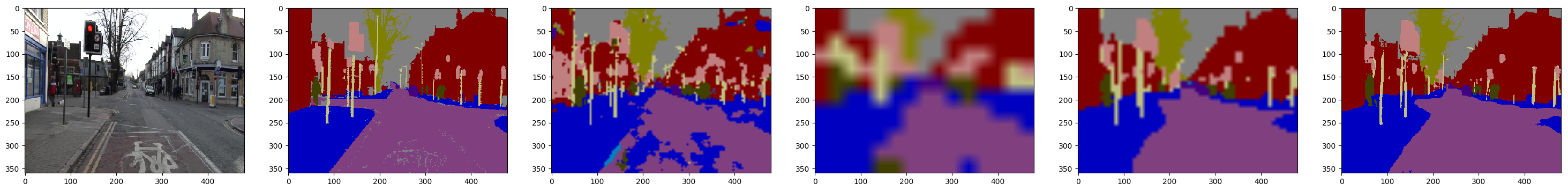}
    }
    
    \subfloat{
    \includegraphics[width=\linewidth]{img_30.png}
    }
    
    \subfloat{
    \includegraphics[width=\linewidth]{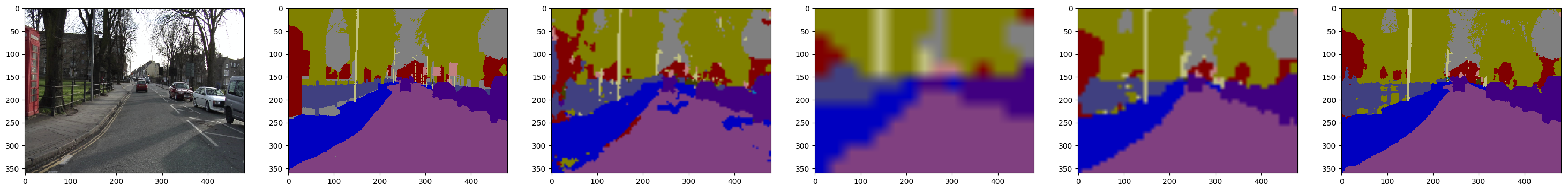}
    }
    
    \subfloat{
    \includegraphics[width=\linewidth]{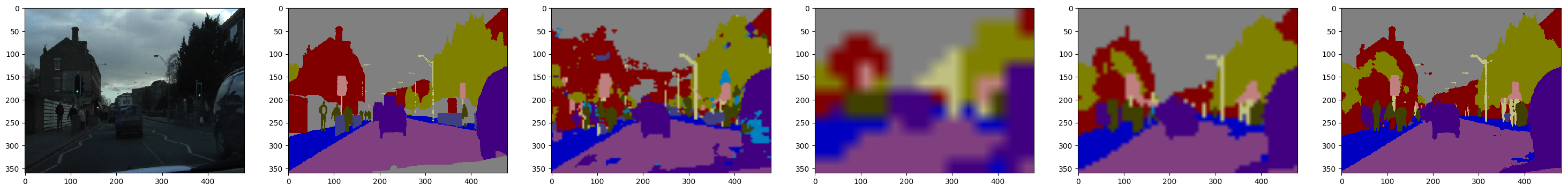}
    }
    
    \subfloat{
    \includegraphics[width=\linewidth]{img_120.png}
    }
    
    \subfloat{
    \includegraphics[width=\linewidth]{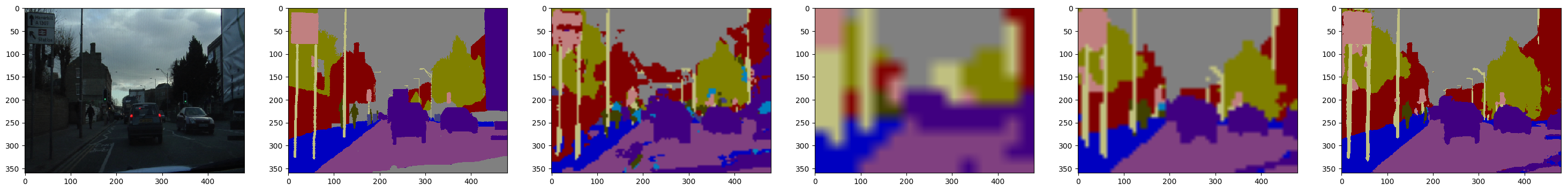}
    }
    
    \subfloat{
    \includegraphics[width=\linewidth]{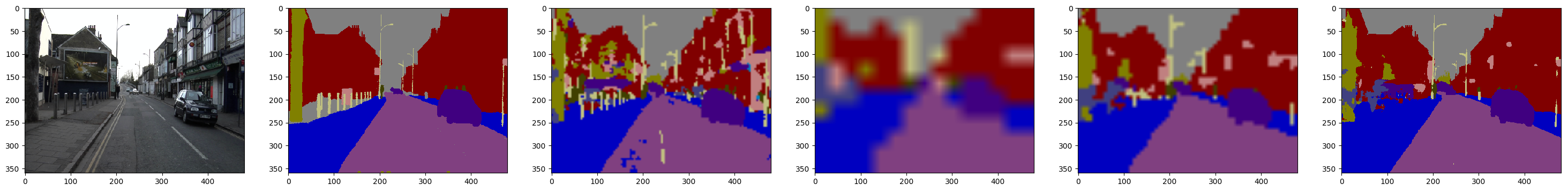}
    }
    
    \subfloat{
    \includegraphics[width=\linewidth]{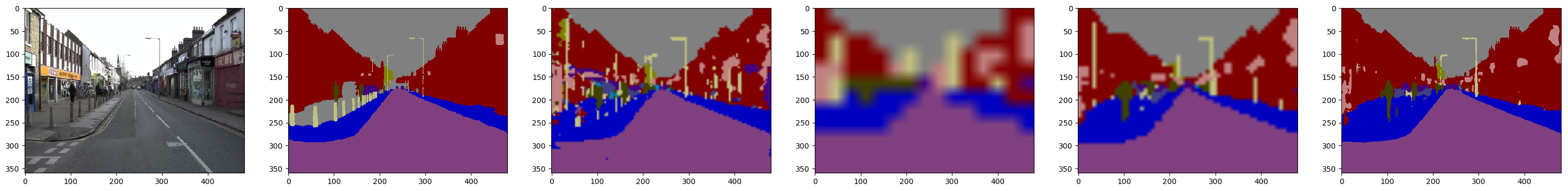}
    }

    \caption{Each row: input image, ground truth labeling, and any scene parsing results at 1/4, 1/2, 3/4 and the final layer.}
    \label{fig:scene_parsing_appendix}
\end{figure}

We show additional semantic segmentation results in Figure~\ref{fig:scene_parsing_appendix}. We also note in Figure~\ref{fig:fc_computation} how the computation is distributed through the 11 blocks in FC-DenseNets and FC-\logdenses. In particular, more than half of the computation is from the final two blocks because the final blocks have high resolutions, making them exponentially more expensive than layers in the mid depths and final layers of image classification networks.

\end{document}